\documentclass[a4paper,conference]{IEEEtran}
\usepackage{cite}
\usepackage{multirow}

\ifCLASSINFOpdf
   \usepackage[pdftex]{graphicx}
   \DeclareGraphicsExtensions{.pdf}
\else
\fi
\usepackage{amsmath}
\usepackage{algorithmic}
\usepackage{array}
\begin{document}
\title{Visual Localization of Key Positions for Visually Impaired People}

% \author{\IEEEauthorblockN{Ruiqi Cheng, Kaiwei Wang and Shufei Lin}
% \IEEEauthorblockA{College of Optical Science and Engineering\\
% Zhejiang University\\
% Hangzhou, China\\
% Email: {rickycheng, wangkaiwei}@zju.edu.cn}
% \and
% \IEEEauthorblockN{Longqing Lin}
% \IEEEauthorblockA{Kr Vision Technology Co., Ltd.\\
% Hangzhou, China\\
% Email: longqing.lin@krvision.cn}
% \and
% \IEEEauthorblockN{Kailun Yang}
% \IEEEauthorblockA{College of Optical Science and Engineering\\
% Zhejiang University\\
% Hangzhou, China\\
% Email: elnino@zju.edu.cn}}


\author{\IEEEauthorblockN{Ruiqi Cheng\IEEEauthorrefmark{1},
Kaiwei Wang\IEEEauthorrefmark{1},
%Shufei Lin\IEEEauthorrefmark{1},
Longqing Lin\IEEEauthorrefmark{2} and
Kailun Yang\IEEEauthorrefmark{1}}
\IEEEauthorblockA{\IEEEauthorrefmark{1}College of Optical Science and Engineering, Zhejiang University, Hangzhou, China\\
Email: \{rickycheng, wangkaiwei, elnino\}@zju.edu.cn}
\IEEEauthorblockA{\IEEEauthorrefmark{2}Kr Vision Technology Co., Ltd., Hangzhou, China\\
Email: longqing.lin@krvision.cn}}


\maketitle

\begin{abstract}
On the off-the-shelf navigational assistance devices, the localization precision is limited to the signal error of global navigation satellite system (GNSS). During travelling outdoors, the inaccurately localization perplexes visually impaired people, especially at key positions, such as gates, bus stations or intersections. The visual localization is a feasible approach to improving the positioning precision of assistive devices. Using multiple image descriptors, the paper proposes a robust and efficient visual localization algorithm, which takes advantage of priori GNSS signals and multi-modal images to achieve the accurate localization of key positions. In the experiments, we implement the approach on the wearable system and test the performance of visual localization under practical scenarios.
\end{abstract}


% \fi
\IEEEpeerreviewmaketitle

\section{Introduction}
In the world, around 253 million people live with vision impairments~\cite{WHO2017}. Navigational assistance is one of crucial demands for visually impaired people in their daily life. Thanks to the booming smart phones and mobile Internet, visually impaired people are easy to get access to the off-the-shelf navigational applications, which rely merely on the GNSS (Global Navigation Satellite System) signal. Generally, the localization error of GNSS is more than several meters under ordinary conditions, and is up to dozens of meters under severe weather conditions. In terms of the off-the-shelf navigational assistance, the positioning precision is insufficient. Imagining a person with visual impairments standing at the vicinity of a turning, it is tough for her or him to locate where should be the exact position to take a turn relying merely on GNSS localization. Therefore, applying a localization approach with less error to the practical navigational assistance is of vital importance for alleviating the underlying hazards caused by inaccurate positioning.

There were different sorts of non-visual localization utilized in the assistive navigation for the visually impaired, such as RFID (Radio Frequency IDentification) technology~\cite{RFID} or WiFi localization~\cite{WIFI}. Nevertheless, plenty of RFID or WiFi tags have to be deployed at key positions to achieve the ubiquitous navigation. In contrast, visual localization based on GNSS signal is a more feasible approach.

\begin{figure}[!t]
\centering
\includegraphics[width=\linewidth]{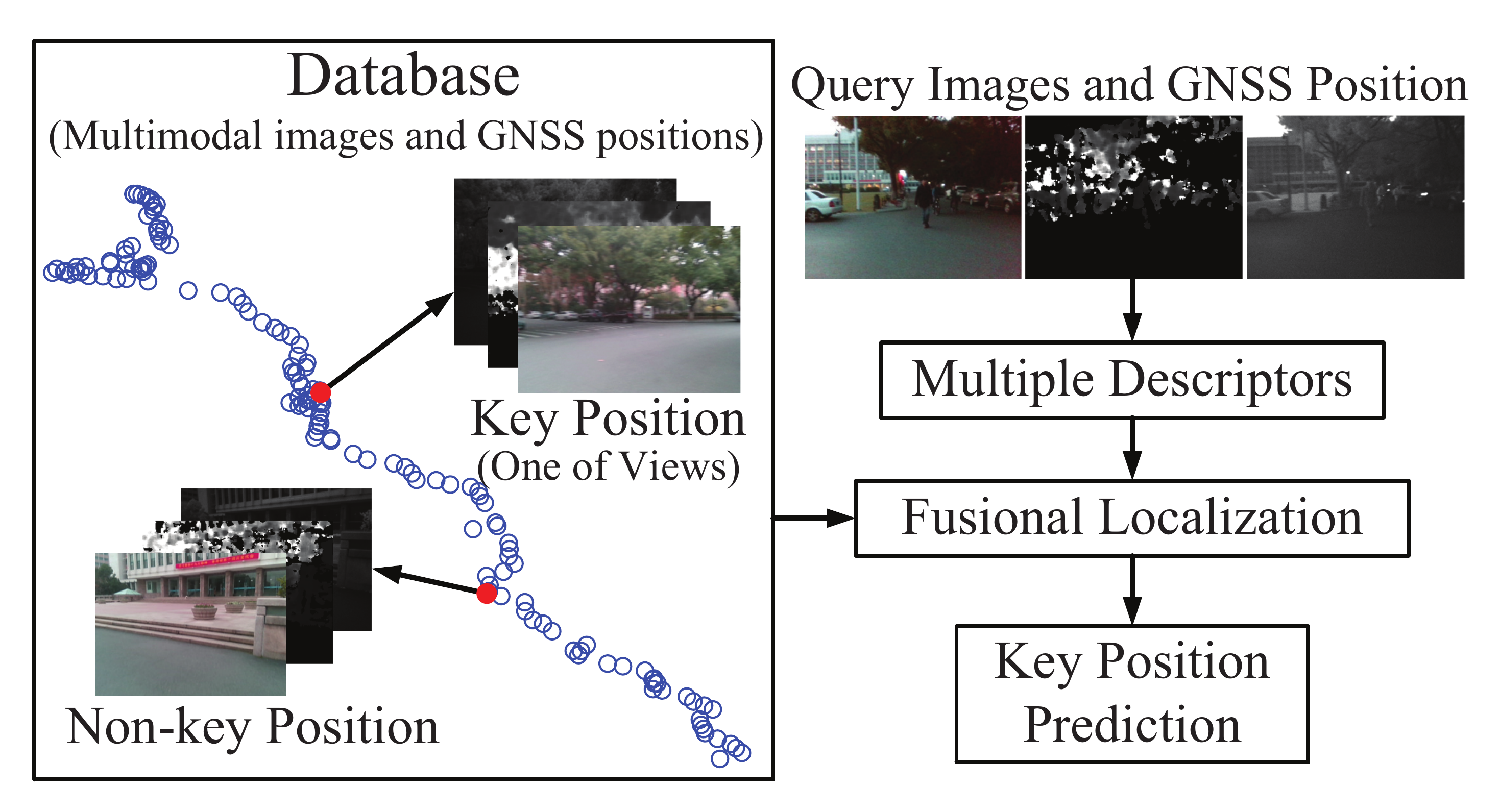}
\caption{General block diagram of the proposed framework for visual localization of key positions.}
\label{fig_sim}
\end{figure}

In this paper, we propose a robust and real-time visual localization approach based on the multiple descriptors of multi-modal images and the GNSS signal. Shown as Fig.~\ref{fig_sim}, the proposed key position localization approach involves a priori database containing images and GNSS positions, query images and GNSS positions, as well as key position prediction fusing image with GNSS. As a coarse localization, GNSS signals are utilized to shrink the retrieval range of images. Visual localization is arguably a fine-tuned localization, which promotes localization precision by retrieving best matching query images from database.

Composed of GNSS signals and multiple descriptors, a traversed trajectory is recorded in the database, where key positions and non-key positions are both included. The GNSS signal consists of the coordinate of longitude and latitude. Key positions denote the positions which the visually impaired user attaches the importance to, such as turnings, gates, barriers, bus stations, as well as trees on the roads, etc. At the key positions, the user is required to turn around, so as to capture scenes with different views in addition to the frontal view. The non-key positions are the rest positions in the traversed trajectory, where only frontal view images are recorded. 

When the visually impaired user traverses the recorded trajectory again, the multiple image descriptors of the query images as well as the GNSS position are fused together to find the nearest positions from the database. Then, the algorithm draws the conclusion whether the current position is a pre-labeled key position.

Due to the changing pose of the wearable device, the captured images, including both database images and query images, are not as stable as those captured from the camera mounted on robots or vehicles. In assistive navigation, the difficulty of visual localization lies in the pose variance between database images and query images, so the multiple descriptors extracted from multi-modal images are utilized to depict scenes better. Different from the visual localization or place recognition in the application of intelligent vehicles, there is no public dataset suitable for the blind assistance. Therefore, the main contributions of our work are fourfold:
\begin{itemize}
\item A real-world visual localization on a wearable devices specially designed for people with impaired vision.
\item Higher localization accuracy compared with GNSS positioning approaches.
\item Real-time response on the limited wearable platform that qualifies the navigational assistance.
\item A challenging dataset containing multi-modal images and GNSS signal for visual localization in the area of blind assistance.
\end{itemize}

\section{Related Work}
The community of assistive navigation achieved different localization solutions, which are applied to various scenarios of navigational assistance. In order to inform the visually impaired with ambient objects, Mekhalfi et al.~\cite{CS} proposed a multi-label scene recognition algorithm based on compressive sensing, which achieves considerable localization accuracy for the scenes once visited. Nevertheless, the compressive sensing has high computational complexity, hence it is not suitable for real-time application. Nguyen and Tran~\cite{GISTBLIND} proposed a scene description based on GIST features and kNN (k-Nearest Neighbor), which classifies query images into key scenes that are visually different. However, the number of classes is too small to enable precise localization. Moreover, appearance variance is not considered in capturing training and testing data, so that robustness is not guaranteed. Ivanov~\cite{indoor} proposed an indoor navigation system, which requires building floor plans. Fusco et al.~\cite{selfLocalize} proposed a self-localization approach based on a street-view panorama and an aerial image. The algorithm locates the visually impaired by extracting crosswalk stripes from both of images, so it is only valid at street intersections. Therefore, to the best of our knowledge, none of visual localization solutions aims or manages to promote the navigational precision outdoors.

The community of intelligent transportation achieved different kinds of visual localization algorithms, which are also called topological localization or place recognition. For autonomous vehicles, Arroyo et al.~\cite{OpenABLE} implemented a visual localization algorithm which is based on LDB (Local Difference Binary) descriptor. However, the algorithm is sensitive to the pose and FOV (Field Of View) of camera, which is hard to be maintained stable on wearable devices. Tian et al.~\cite{Tian_2017_CVPR} proposed a visual localization approach based on GNSS information and bird’s eye view images. Obviously, the bird’s eye view images are not available in assistive navigation.

The evaluation of the visual localization algorithms relies on public datasets, such as KITTI~\cite{KITTI} and CMU CVG VL~\cite{CMU} etc. Unfortunately, these datasets are not designed for assistive navigation. In our cases, the constantly varying camera pose, which is caused by user movement, results in motion blur and diverse FOVs among the database and query images.
\section{Visual Localization}
In order to achieve a satisfactory localization performance, it is of vital importance to depict the scenes comprehensively. For the sake of it, not only are the multi-modal images captured to enrich the input information, but also both holistic and local descriptors are utilized to represent the images. The multi-modal images used in this paper are RGB images, infrared images and depth images. Representing the holistic scene, GIST descriptor is used to alleviate the impact of scene changes, e.g. people, cars or luminance etc. Besides, BoW (Bag of Words) based on the local descriptor is used complementarily for describing image details. As a feature that performs well in multi-modal images, LDB is used to promote the matching robustness by synthesizing multi-modal images. 
\subsection{GIST Descriptor}
\begin{figure}[tb]
\centering
\includegraphics[width=0.8\linewidth]{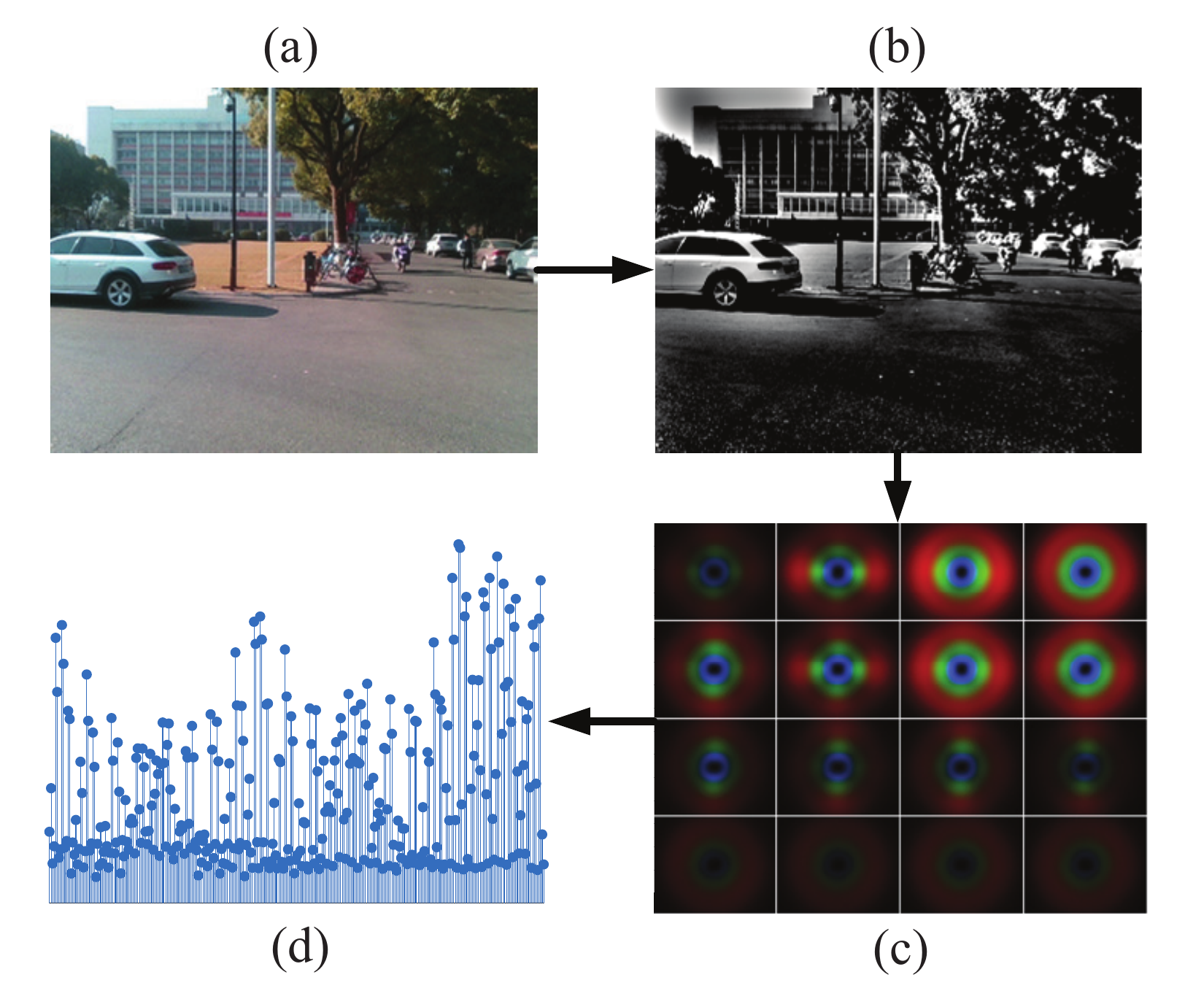}
\caption{GIST descriptor extraction procedures. (a) Input image. (b) Normalized image. (c) Visualization of down-sampled Garbor responses, where different colors denote different scales. (d) GIST descriptor.}
\label{fig_gist}
\end{figure}

Instead of depicting image by local key points, GIST descriptor~\cite{Oliva2001, context} serves as a holistic representation of the scene. As shown in Fig.~\ref{fig_gist}, the extraction procedures of GIST descriptor involve image normalization, Gabor filtering and response averaging. First of all, the image variance caused by different illuminations is ameliorated by applying whitening and local contrast normalization to original images [see Fig.~\ref{fig_gist} (a) and (b)]. Then, Gabor responses are obtained and down-sampled by using several Gabor filters with different orientations and scales [see Fig.~\ref{fig_gist} (c)]. Finally, the responses are  concatenated together to generate the GIST descriptor [see Fig.~\ref{fig_gist} (d)]. In the paper, we use up to 20 Gabor filters belonging to 3 different scales, and the quantity of orientations in each scale (from large to small) is 8, 8 and 4 respectively. The number of blocks is set to 16, which means each Gabor response image is down-sampled to 4$\times$4. Thereby, by merging all of the responses together, the size of final GIST descriptor is up to 320.
\subsection{LDB Descriptor}
\begin{figure}[tb]
\centering
\includegraphics[width=\linewidth]{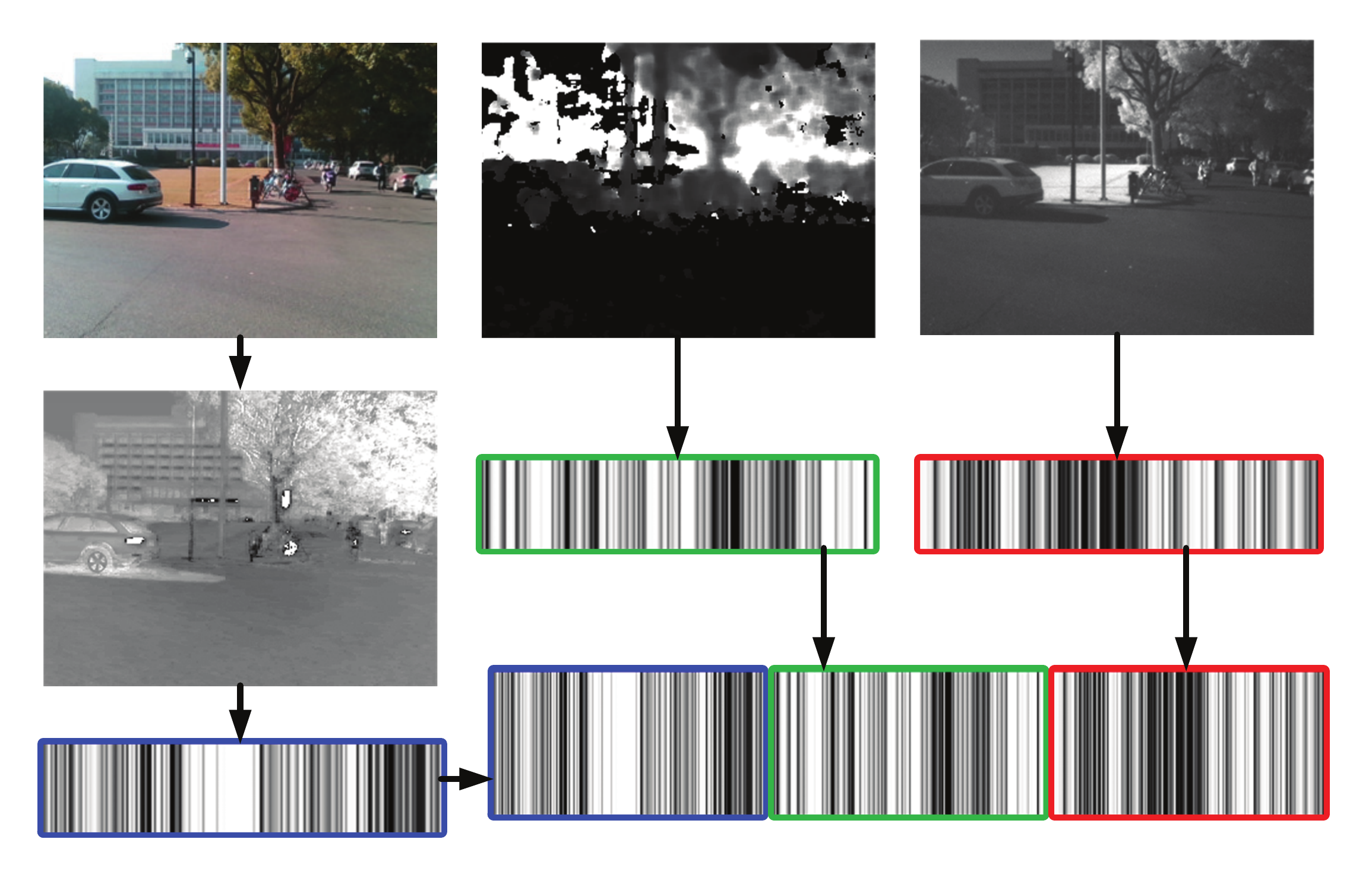}
\caption{LDB descriptor extraction procedures. The input images are composed of RGB, depth and infrared images (from left to right in the first row). Illumination invariance transformation is applied to RGB image, then LDB descriptors are extracted from each image and concatenated together to form the final LDB descriptor.}
\label{fig_ldb}
\end{figure}

Yang and Cheng~\cite{LDB} proposed LDB descriptor, which is highly efficient and sufficiently distinctive. Arroyo et al.~\cite{OpenABLE} utilized concatenated LDB descriptor which is extracted from multi-modal images (RGB image, gradient image and disparity image) to achieve visual localization. In our work, three LDB descriptors are extracted from the RGB, depth and infrared images respectively, and are concatenated to form a compounded LDB descriptor, shown as Fig.~\ref{fig_ldb}. Before extracting LDB descriptor from RGB images, illumination invariance transformation is applied to improve the robustness under the condition of changing lighting~\cite{OpenABLE}.
\subsection{BoW Descriptor}
Originated from text analysis, BoW is widely applied in object and scene categorization, due to its simplicity, computational efficiency and invariance to affine transformation~\cite{Bow2004}. Galvez-López and Tardos~\cite{BinaryWords} proposed a BoW-based place recognition approach, which is widely used in the loop closure of simultaneous localization and mapping. Different from holistic GIST descriptor, BoW descriptor, arguably a further abstraction of local features, represents the scene details to some extent.

\begin{figure}[tb]
\centering
\includegraphics[width=0.85\linewidth]{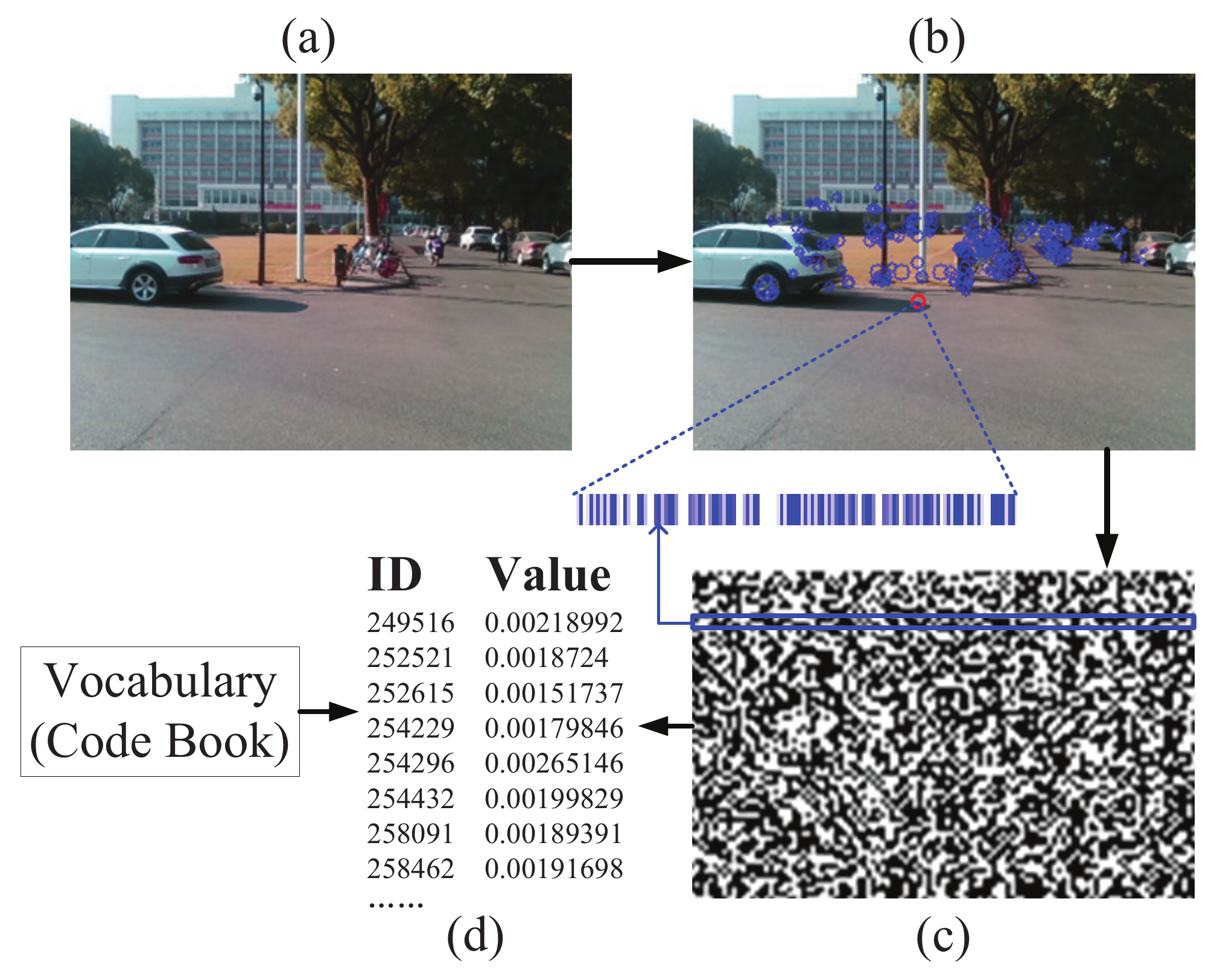}
\caption{BoW descriptor extraction procedures. (a) Input image. (b) Images with FAST key points (blue circles) and rotated BRIEF feature of one key point (blue barcode). (c) ORB features of input images. (d) BoW vector of input image with word ID and its value.}
\label{fig_bow}
\end{figure}

To balance computational efficiency and affine invariance, ORB (Oriented FAST and Rotated BRIEF)~\cite{ORB} is chosen as the local feature of BoW. The key points are extracted by FAST (Features from Accelerated Segment Test), and described by rotated BRIEF (Binary Robust Independent Elementary Features) [see Fig.~\ref{fig_bow} (b)]. The ORB descriptors of all key points are merged together and compose the concatenated descriptors [see Fig.~\ref{fig_bow} (c)]. Subsequently, the BoW descriptor [see Fig.~\ref{fig_bow} (d)] is generated using the ORB descriptor and pre-trained vocabulary~\cite{DBoW3}, in which visual words are clustered to form a hierarchical tree using massive ORB descriptors of training images. In view that the ORB vocabulary used in the paper is trained by gray-scale image (derived from the RGB image) and rare key points are extracted in depth and infrared image, BoW descriptor is only applied to RGB images.
\subsection{Key Position Prediction}
\begin{figure}[tb]
\centering
\includegraphics[width=\linewidth]{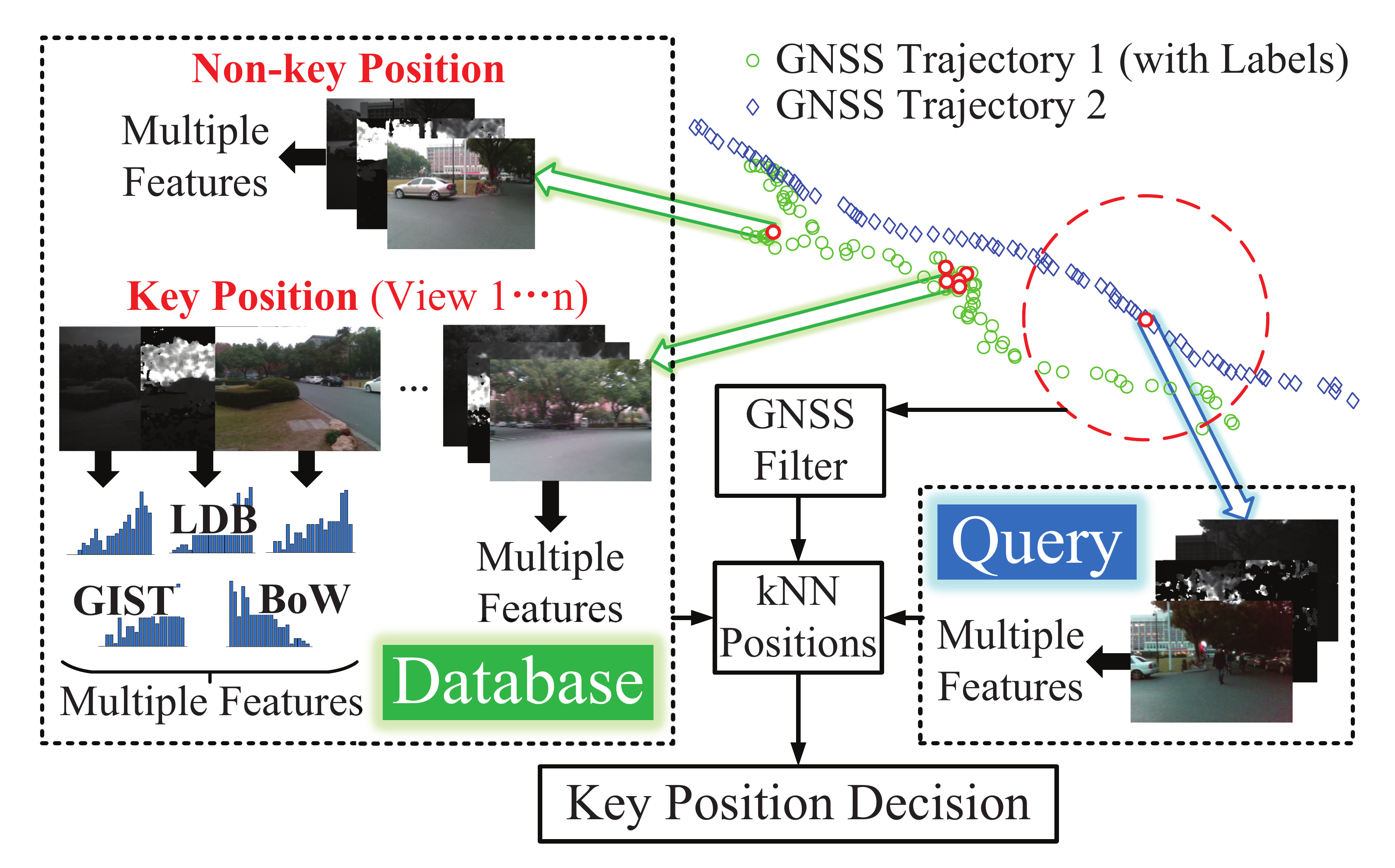}
\caption{The flow chart of key position prediction using the multiple image descriptors and GNSS signals.}
\label{fig_keyPos}
\end{figure}

When traversing a trajectory at the first time, the multiple image descriptors (GIST, LDB and BoW) with corresponding GNSS signals are recorded as database (e.g. trajectory 1 in Fig.~\ref{fig_keyPos}), where different views of images at a key positon are captured as many as possible. During traversing the trajectory again (e.g. trajectory 2 in Fig.~\ref{fig_keyPos}), kNN algorithm is used to find the matching results from the database based on the query image descriptor and GNSS signal, thus the key position localization is decided by the labels of the matching images. Serving as a coarse localization, the query GNSS coordinate is utilized to constrain the range of image retrieval. Only database images whose GNSS coordinate is within the range $r$ of the query GNSS coordinate (e.g. the green points within the red circle in Fig.~\ref{fig_keyPos}) are taken as candidates to be matched.

The similarity of the query image and the filtered database images is measured by different distances of images descriptors. Euclidean distance and Hamming distance~\cite{OpenABLE} are applied to match GIST descriptors and LDB descriptors respectively. The distances between BoW descriptors are measured with L$_1$-score~\cite{BinaryWords}. The dimensions of GIST and LDB descriptor are limited, so brute force matching is utilized to find the $k_{GIST}$ nearest neighbors of GIST and $k_{LDB}$ nearest neighbors of LDB. However, the dimension of BoW is large, hence inverse index of vocabulary tree is used to quickly access the $k_{BoW}$ nearest neighbors~\cite{BinaryWords}. The kNN matching results of the query image are attained by incorporating all nearest neighbors of GIST, LDB and BoW descriptor. The number of matching images selected from the database is up to $k=k_{GIST} + k_{BoW} + k_{LDB}$. If the number of images that are labeled as key position out of matching images exceeds the threshold $n$, the query position is predicted as a key position.
\section{Experiments and Discussions}
\begin{figure}[tb]
	\centering
	\includegraphics[width=0.65\linewidth]{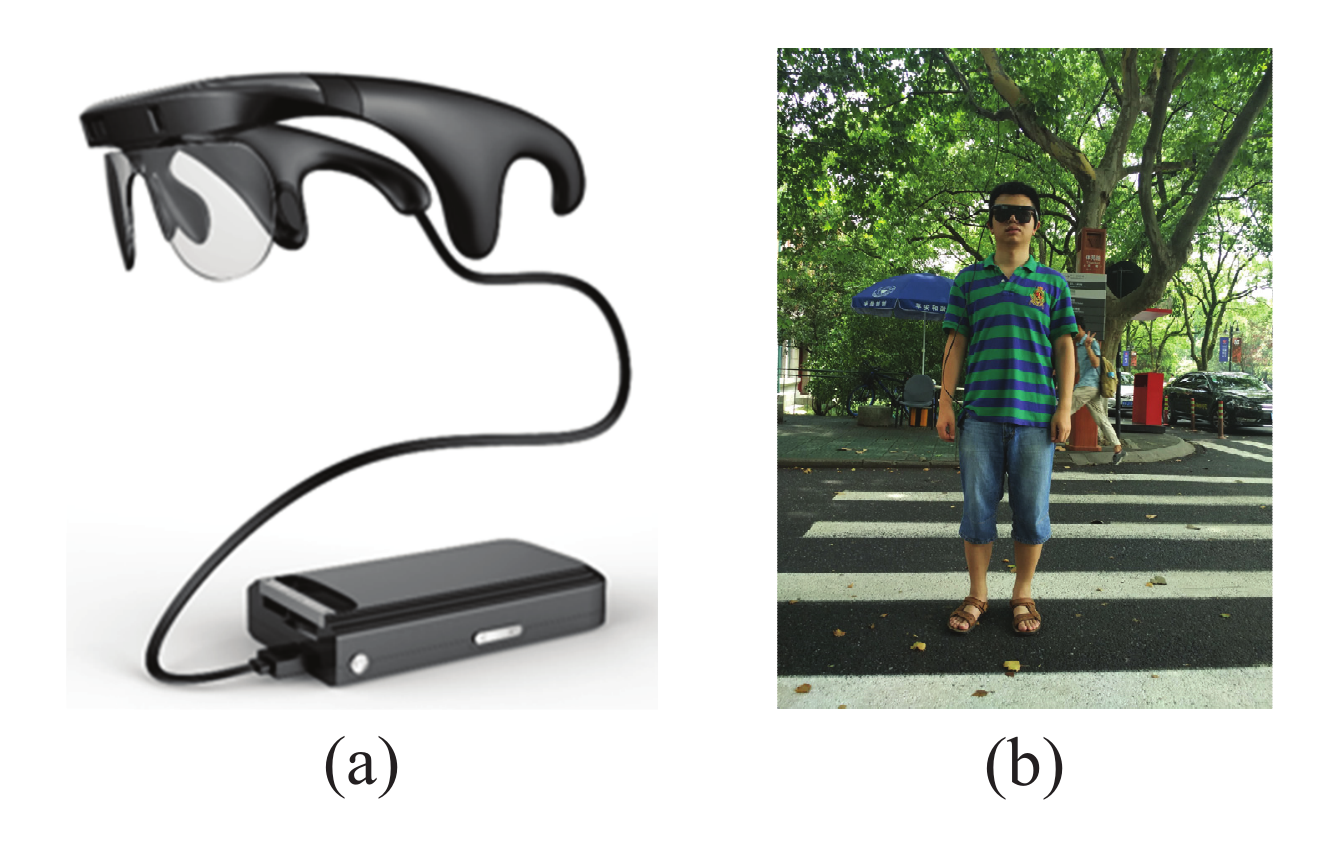}
	\caption{(a) Intoer: the wearable navigational devices for visually impaired people. (b) A user is wearing Intoer.}
	\label{system}
\end{figure}

In order to achieve navigational assistance for people with visual impairments, we developed a wearable assistive system Intoer~\cite{KrVision}, which is comprised of the multi-modal camera RealSense~\cite{Intel}, a customized portable processor with GNSS module, and a pair of bone-conduction earphones~\cite{AfterShokz}, as shown in Fig.~\ref{system}. Based on the system, we have achieved various assistive utilities, including traversable area and hazard awareness~\cite{s17081890}, crosswalks and traffic lights detection~\cite{CW,Cheng2017} etc. Using the wearable system Intoer, we capture real-world scenes to build a challenging dataset, which consists of a series of time-ordered multi-modal images and GNSS signals~\cite{VLdataset}. The capture interval of two successive frames is one second. The resolutions of multi-modal images are set to 320$\times$240. The real-world dataset features illumination and pose variances between database images and query images. 
%\begin{table}[tbp]
%	\renewcommand{\arraystretch}{1.3}
%	\caption{Statistics of Visual Localization Dataset}
%	\begin{center}
%		\begin{tabular}{|c|c|c|c|c|c|c|}
%			\hline
%			\textbf{Route}&\textbf{Length}&\textbf{Trajectory}& \textbf{Size}&\multicolumn{3}{c|}{\textbf{Key Position}}\\
%			\hline
%			\multirow{3}{*}{1}&\multirow{3}{*}{150m}&database&142&13&11 &{}\\
%			\cline{3-7}
%			{}&{}&query 1&105&{}&{} &{}\\
%			\cline{3-7}
%			{}&{}&query 2&97&{}&{} &{}\\
%			\hline		
%		\end{tabular}
%		\label{table_dataset}
%	\end{center}
%\end{table}
\subsection{Localization Precision of Different Descriptors}
The index difference between the matching result and the ground truth reflects the localization precision, since the database is recorded in time order and the successive data are with adjacent indexes. For each image descriptor, the minimal index difference among the nearest neighbors quantifies the performance of that descriptor. As a criterion of descriptor performance, full trajectory error is the mean value of minimal index differences throughout the full trajectory. Another criterion is sensitivity, which is the ratio of matched query images number to all query images number. Thereby, in order to validate the adopted image descriptors (GIST, LDB and BoW), we test those descriptors on the our dataset and compare them with CS (Compressive Sensing). Transforming high-dimensional images into low-dimensional descriptors, CS achieves sparse representation of the original image. In this paper, we use CoSaMP (Compressive Sampling Matching Pursuit)~\cite{CoSaMP} to extract the sparse descriptor of image.

The comparison experiment is carried out on a 150-meter long route of the dataset, which comprised of a database trajectory with 142 multi-modal images and GNSS signals and two query trajectories. TABLE~\ref{table_descriptor} gives the full trajectory error and sensitivity of different descriptors on different multi-modal image combinations, when $k_{GIST}=k_{BoW}=k_{LDB}=5$ and $r=0.02$. Local descriptor BoW on RGB image achieves the best performance among all descriptors, in view of the lowest error and the highest sensitivity. Meanwhile, holistic GIST descriptor on RGB images has lower error and considerable sensitivity compared with GIST extracted from other images. Both GIST and BoW are vulnerable to changing illumination, while LDB descriptor performs stable under the circumstance, hence LDB that extracted from RGB-IR-D images is also chosen in this paper. CS descriptor is not considered in this paper, because of its highly computational complexity and low robustness.
\begin{table*}[!t]
\renewcommand{\arraystretch}{1.3}
 %if using array.sty, it might be a good idea to tweak the value of
 %\extrarowheight as needed to properly center the text within the cells
\caption{Performance of Different Image Descriptors on Multi-modal Images}
\label{table_descriptor}
\centering
% Some packages, such as MDW tools, offer better commands for making tables
% than the plain LaTeX2e tabular which is used here.
	\begin{tabular}{|c|c|c|c|c|c|c|c|c|c|c|}
	\hline
	\multicolumn{2}{|c|}{\multirow{2}{*}{\textbf{Trajectory}}}&\multicolumn{4}{c|}{\textbf{RGB}} &\multicolumn{2}{c|}{\textbf{RGB-IR}}&\multicolumn{2}{c|}{\textbf{RGB-IR-D}}&\multirow{2}{*}{\textbf{GNSS}} \\
	\cline{3-10} 
	\multicolumn{2}{|c|}{} & \textbf{{\textit{GIST}}} & \textbf{\textit{BoW}}& \textbf{\textit{CS}}& \textbf{\textit{LDB}}&  \textbf{{\textit{GIST}}}& \textbf{\textit{LDB}}&  \textbf{{\textit{GIST}}}& \textbf{\textit{LDB}}&{} \\
	\hline
	\multirow{2}{*}{1$^{\mathrm{a}}$}	&Sensitivity&	\textbf{89.6\%}&	\textbf{96.2\%}	&94.0\%&	91.5\%&	69.8\%&	90.6\%&	54.7\%&	\textbf{93.4\%}&	100.0\%\\
	\cline{2-11} 
	&	Error&	\textbf{7.73}&	\textbf{5.08}&	9.55&	9.03&	9.72&	9.35&	13.03&	\textbf{9.20}&	9.11  \\
	\hline
	\multirow{2}{*}{2$^{\mathrm{b}}$}	&Sensitivity&\textbf{75.3\%}&	\textbf{99.0\%}&	87.1\%	&88.7\%	&89.7\%	&90.7\%&	94.9\%&	\textbf{82.5\%}	&100.0\%\\
	\cline{2-11}  
	&	Error & \textbf{4.33} &\textbf{2.21}&	8.88 & 9.43 & 5.62 &		8.31 &		5.07 &		\textbf{9.10} &		3.74 \\
	\hline	
	\multicolumn{11}{r}{$^{\mathrm{a}}$105 query data with illumination different from database.~$^{\mathrm{b}}$97 query data with illumination similar to database.}
	\end{tabular}
\end{table*}
\subsection{Performance of Key Position Prediction}
Using the experimentally validated image descriptor configuration, we test the precision and recall of key position prediction on a 400-meter route of the dataset, where three query trajectories correspond to one database trajectory featuring 291 multi-modal images and GNSS signals. According to the key position prediction results of all query images, we count the total number of true positives, false positives and true negatives, and define them respectively as TP, FP and FN. Herein, precision and recall are defined as:
\begin{equation}
\centering
Precision=\frac{TP}{TP+FP}
\label{e1}
\end{equation}
\begin{equation}
\centering
Recall=\frac{TP}{TP+FN}
\label{e2}
\end{equation}

\begin{figure}[!t]
	\centering
	\includegraphics[width=\linewidth]{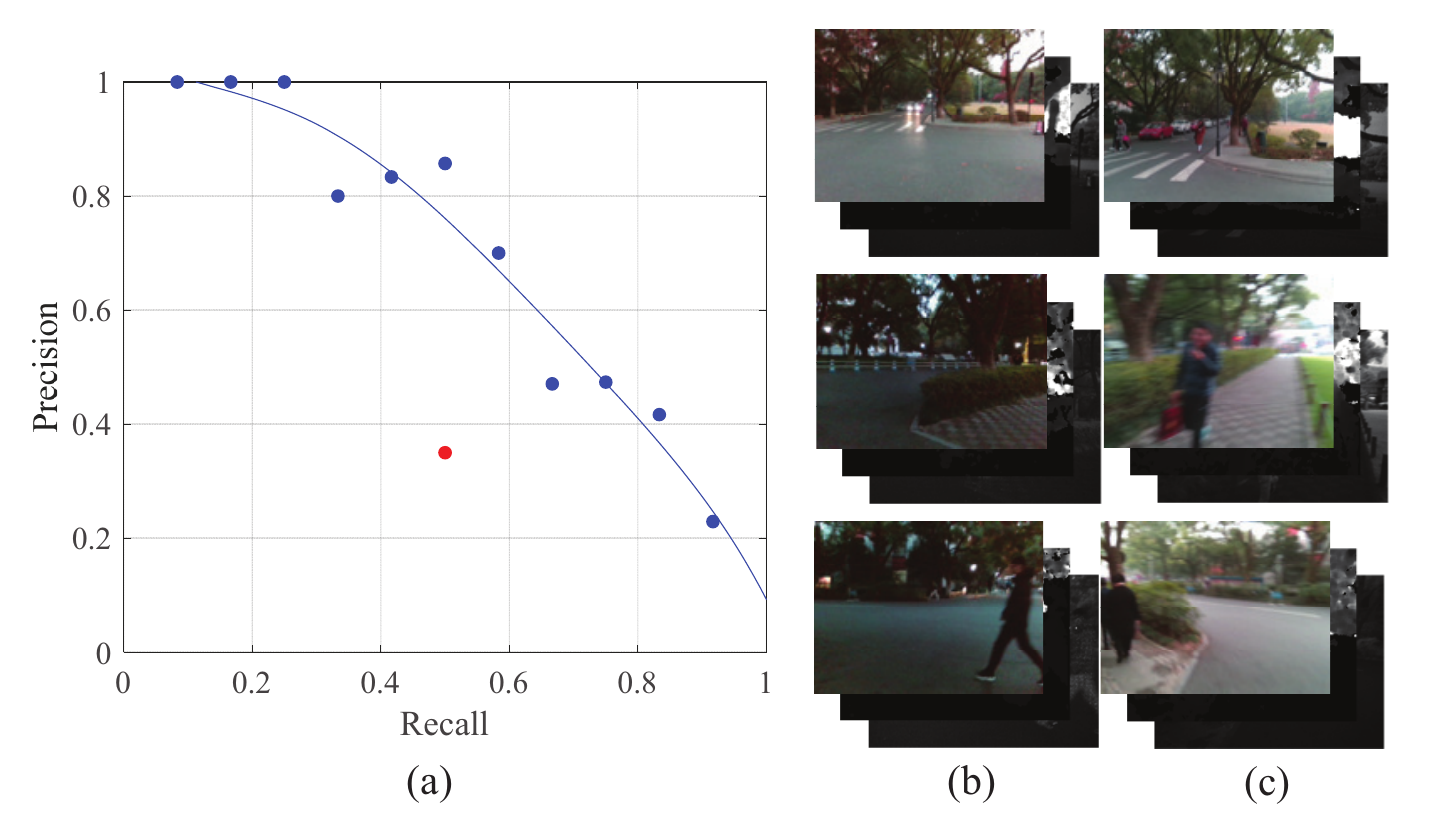}
	\caption{(a) The precision-recall curve of visual localization (blue), the precision and recall of GNSS localization (red). (b) The labeled key positions in database trajectory with 17, 17 and 15 data per key position respectively. (c) The matching positions in query trajectory.}
	\label{fig_roc}
\end{figure}

On one of query trajectories, a grid search is executed to find the optimal values of parameters, including the number of nearest neighbors ($k_{GIST}$, $k_{LDB}$ and $k_{BoW}$), GNSS range~$r$ and key position threshold~$n$. In Fig.~\ref{fig_roc}~(a), the different precisions and recalls derived from different parameter values are utilized to fit the precision-recall curve of visual localization. Compared with mere GNSS localization, the visual localization attains better recall and precision. Setting parameters to the optimal values, we test key position prediction again on the same query trajectory. Fig.~\ref{fig_roc}~(b) and (c) illustrate that the predicted key positions are close to the ground truths, even illumination and FOV differ.

In order to validate robustness, we test key position localization on the rest two query trajectories of the dataset, which are traversed reversely to the database trajectory. Recall and key position localization error (the mean index difference between predicted positions and ground truths) are used to characterize the performance. As presented in TABLE~\ref{table_performance}, the proposed localization approach achieves higher recall and lower error compared with GNSS localization.
\begin{table}[tbp]
	\renewcommand{\arraystretch}{1.3}
	\caption{Performance of Visual Localization}
	\begin{center}
		\begin{tabular}{|c|c|c|c|c|c|}
			\hline
			\multirow{2}{*}{\textbf{Trajectory}}& \textbf{Image}&\multicolumn{2}{c|}{\textbf{Visual Localization}} &\multicolumn{2}{c|}{\textbf{GNSS}} \\
			\cline{3-6} 
			{} & \textbf{Number}& \textbf{\textit{Error}}& \textbf{\textit{Recall}}&  \textbf{\textit{Error}}& \textbf{\textit{Recall}} \\
			\hline
			1$^{\mathrm{a}}$&215&\textbf{11.5} &\textbf{36.4\%}&14.8&0  \\
			\hline
			2$^{\mathrm{a}}$&208&\textbf{4.1} &\textbf{29.4\%}&8.4&0  \\
			\hline
			\multicolumn{6}{r}{$^{\mathrm{a}}$Walking along the database trajectory reversely.}
		\end{tabular}
		\label{table_performance}
	\end{center}
\end{table}
\subsection{Field Tests}
\begin{figure*}[!t]
	\centering
	\includegraphics[width=0.65\linewidth]{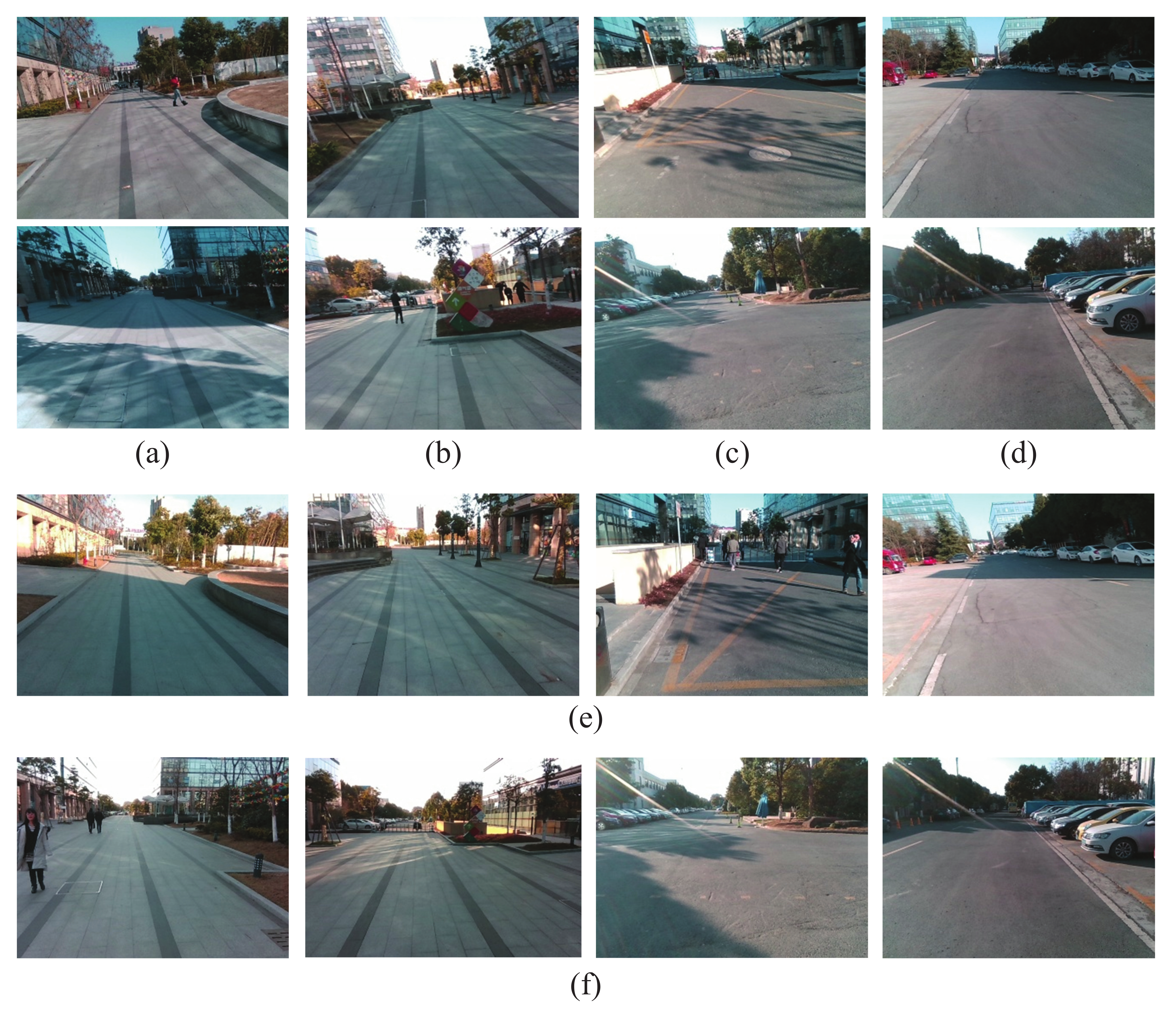}
	\caption{Labeled and predicted key positions in the field test carried out on a 400-meter route. In (a) - (d), four labeled key positions, each of which contains different views of images, are presented (depth and infrared images are omitted), respectively. (e) and (f) present the key position predictions of a round trip on the route.}
	\label{fig_ft}
\end{figure*}

The wearable system Intoer is utilized to run the proposed algorithm to validate its key position localization performance in practical scenarios. The experiment is carried out on an approximately 400-meter route from point A to point B. When we traverse the routes at first time, four positions are labeled as key positions, where different views of the scenes are captured by the multi-modal camera, meanwhile the GNSS signal is recorded. Two views of each labeled key position are presented in Fig.~\ref{fig_ft} (a)-(d). When the labeled routes are traversed again, the algorithm predicts whether the current position is a key position according to the acquired multi-modal images and GNSS signals. Fig.~\ref{fig_ft} (e) presents the predicted key positions when the user walks from point A to point B, and Fig.~\ref{fig_ft} (f) presents the predicted key positions when the user walks in the reverse direction of the route. The field test illustrates that the proposed approach achieves accurate localization in practical use. As for efficiency, the field test verifies that the proposed algorithm implemented on the portable platform achieves real-time response. The processing time to predict one position is around 250 ms, which is less than the interval of image capture (one second in this paper).

\section{Conclusion}
In order to achieve navigational assistance for people with visual impairments, we propose a visual localization approach which combines GNSS signals with multiple image descriptors extracted from multi-modal images.

For high localization precision, multi-modal images, RGB, depth and infrared images included, are utilized in the paper. Besides, multiple descriptors, both holistic and local, are used to depict image comprehensively. Combined with GNSS signals, the approach matches the query images with the database images, and predicts whether the current location is a pre-labeled key position. The experiments carried on both the self-captured dataset and practical scenarios indicate that the proposed visual localization approach is accurate and efficient for the assistive navigation of visually impaired people.

In the future, the visual localization will be improved on the environmental robustness in terms of sequential localization and low illuminance localization.


%\section*{Acknowledgment}




\bibliographystyle{IEEEtran}
\bibliography{IEEEabrv,mybibfile}

%\begin{equation}
%\centering
%\ln\frac{y}{1-y}=\mathbf{w^Tx}+b
%\end{equation}
\end{document}